\pdfoutput=1
\documentclass[11pt]{article}
\usepackage[utf8]{inputenc}
\usepackage[T1]{fontenc}
\usepackage{mathptmx}
\usepackage{courier}

\usepackage[margin=2.5cm]{geometry}
\usepackage{booktabs}
\usepackage{graphicx}
\usepackage{amsmath}
\usepackage{microtype}
\usepackage[hidelinks]{hyperref}
\usepackage{siunitx}
\usepackage{caption}
\title{The Invisible Language Tax:\\ Token Premiums of French and Regional Languages in 2026 LLM Tokenizers,\\ and a French-Optimized Prototype}
\author{Thomas Serval\\ Baracoda AI Labs \& NEOMA Business School}
\date{September 2026}

\begin{document}
\maketitle

\begin{abstract}
Large language model (LLM) services are billed per token, and context windows are measured in tokens, yet the number of tokens needed to express the same content varies across languages. We update the measurement of this \emph{token premium} on seven tokenizers of widely used 2026 models (OpenAI o200k, Meta Llama~3, Alibaba Qwen3, DeepSeek V3/V4, Google Gemma~3, Mistral Tekken, and the Claude generation-5 tokenizer measured through Anthropic's counting API), using parallel corpora (NTREX-128, 124 non-English reference translations; the Universal Declaration of Human Rights for regional languages). CroissantLLM's French--English tokenizer serves only as a specialised comparator. For identical content, French requires 31\% to 58\% more tokens than English on these seven tokenizers, whereas Simplified Chinese ranges from 5\% fewer to 40\% more and is cheaper than French on six of seven current tokenizers. Regional and overseas languages of France (Breton, Corsican, Occitan, Picard, Walloon, Tahitian, Creoles) pay roughly 1.6 to 3.3 times the English count. We decompose the French premium into text length and per-character compression, discuss how full-history re-sending, tiered long-context pricing and fixed context windows amplify the absolute gap in agentic use, and report a controlled experiment (BPE tokenizers trained on 48\,MB of Europarl with a 50k vocabulary) in which adding French to the training data quickly reduces the premium, with diminishing returns and a growing cost for English. Finally, we present Baracoda~FR~v1.2, a byte-level BPE prototype with the same vocabulary size as Tekken (131{,}072), built in three measured iterations. On a final test of six corpora never consulted during design, with a protocol declared fixed beforehand, it uses 11.5\% fewer tokens than Tekken on French (95\% CI: $-11.9\%$ to $-11.1\%$) and 3.7\% fewer on English; the result holds after removing test sentences that overlap the training data, and with an ordinary-token budget equal to Tekken's. The Tekken files of current Mistral models (v13, v15) give identical counts. It also uses 2.7\% fewer tokens on Python code, at the cost of 30--38\% more tokens on Spanish and German. At a comparable vocabulary size (\num{32768} vs.\ \num{32000} tokens) it does not outperform CroissantLLM. We state these results as segmentation results only: their effect on model quality and task cost remains to be demonstrated.
\end{abstract}

\section{Introduction}
Commercial LLMs are priced per token and constrained by context windows counted in tokens. Tokenizers, however, are trained on corpora dominated by English, so the same content is split into more pieces in other languages. Petrov et al.~\cite{petrov2023} named this discrepancy the \emph{tokenization premium} and showed premiums of up to 15$\times$ for some languages; Ahia et al.~\cite{ahia2023} linked it to API cost disparities and lower model utility. Both studies predate the tokenizers of 2025--2026.

This note has three goals. First, to update the measurement on the tokenizers actually used in 2026, with a focus on French, the regional and overseas languages of France, and a comparison with Chinese. Second, to make explicit the economic consequences, including in long-context use. Third, to test whether a tokenizer designed for French can reduce the premium, and to report honestly where it does not.

\paragraph{Contributions.}
(1) Token premiums for 124 non-English reference translations of NTREX-128 on six openly available tokenizers, and French and Chinese premiums for the Claude generation-5 tokenizer via its counting API (seven tokenizers in all), with document-level confidence intervals; all local counts are cross-checked against official implementations. CroissantLLM is used only as a specialised French--English comparator (Section~\ref{sec:proto}) and is excluded from the ranges reported for these seven.
(2) Premiums for 18 regional, overseas and Francophone languages on a parallel text.
(3) A decomposition of the French premium and a controlled experiment isolating the effect of the French share in tokenizer training data.
(4) A prototype tokenizer, Baracoda~FR~v1.2, compared with Mistral Tekken and CroissantLLM, including a final test with a protocol declared fixed beforehand and a comparison at comparable vocabulary size in which it does not win.

\section{Related work}
Petrov et al.~\cite{petrov2023} introduced tokenizer parity on FLORES-200 and showed that even closely related European languages can pay about 50\% more than English, and that per-character pricing does not remove the disparity. Ahia et al.~\cite{ahia2023} quantified API cost disparities and showed that over-fragmented languages also tend to obtain worse results. Lundin et al.~\cite{lundin2025} found that higher fertility predicts lower accuracy across 16 African languages. In 2026, Roy et al.~\cite{roy2026} audited technical content in five languages and expressed premiums as lost context capacity, M-GATE~\cite{mgate2026} reported per-character cost multipliers across providers, and a study of prompt compression~\cite{lostcompression2026} found premiums of 1.3--1.8 on nine languages. Another 2026 study showed that Chinese is not inherently more token-efficient~\cite{mythbuster2026}. On the design side, CroissantLLM~\cite{faysse2024} trained a French--English model with a bilingual 32k tokenizer and reported improved French fertility; Parity-aware BPE~\cite{parity2025} modifies merge selection to favour poorly compressed languages; SuperBPE~\cite{superbpe2025} and SupraTok~\cite{supratok2025} learn tokens that cross word boundaries. Vocabulary transfer methods such as FOCUS~\cite{focus2023} and TokAlign~\cite{tokalign2025} address how to adapt a pretrained model to a new tokenizer.

\section{Measuring the token premium}
\paragraph{Corpora.} {\sloppy NTREX-128~\cite{ntrex} contains 1{,}997 English news sentences from 123 documents (\texttt{DOCUMENT\_IDS.tsv}) and 128 professional reference translations: 129 text files in all. Three references are English variants (GB, IN, US), leaving 125 non-English references; we use 124, excluding the second Spanish reference (\texttt{ref-2.spa}) so that each variety appears once. These 124 references are varieties rather than distinct languages: some are variants of one language (e.g.\ French and Canadian French, Portuguese and Brazilian Portuguese, Serbian in two scripts, Chinese simplified and traditional). For regional languages absent from NTREX we use the Universal Declaration of Human Rights (UDHR, \texttt{udhr2}), about 1{,}750 English words, available in some 390 languages. Additional parallel and monolingual test sets from Universal Dependencies~\cite{ud} are used in Section~\ref{sec:proto}.\par}

\paragraph{Tokenizers.} OpenAI o200k (GPT-4o, GPT-5 family, gpt-oss), Meta Llama~3.1, Alibaba Qwen3, DeepSeek V3 and V4 (identical counts on our texts, merged), Google Gemma~3 (same tokenizer as Gemini~2.0~\cite{gemma3}), Mistral Tekken as shipped with \texttt{mistral-common} 1.12 (the Mistral Nemo tokenizer, v3); the official files of Ministral~3 and Mistral Large~3 (Tekken v13), Mistral Medium~3.5 and Mistral Small~4 (Tekken v15, same file) give identical counts on all 15 corpora, differing only in special tokens (checked in Python on a fresh environment). These seven tokenizers support the main measurement; CroissantLLM is added only as a specialised French--English comparator. Local counts were cross-checked token-for-token against official implementations (\texttt{mistral-common}; publishers' Hugging Face repositories for OpenAI, DeepSeek, Qwen3; public mirrors for the gated Llama~3.1 and Gemma~3 repositories). The Claude generation-5 tokenizer (shared by Opus~5.5, Sonnet~5 and Fable~5.1) is not public; it was measured sentence by sentence (7{,}991 requests) with Anthropic's free token-counting endpoint, subtracting a 7-token message overhead. The overhead was measured by difference: seven different texts sent once, twice and three times all give 7 tokens. An earlier draft subtracted 8 tokens, underestimating Claude by one token per sentence; all figures here are corrected, and sentence-level and concatenated counts agree within one token.

\paragraph{Metric.} The premium of language $\ell$ under tokenizer $T$ is the ratio of total tokens for the translated corpus to total tokens for the English source:
\begin{equation}
P_T(\ell)=\frac{\sum_i |T(x_i^{\ell})|}{\sum_i |T(x_i^{\text{en}})|}.
\end{equation}
Texts are NFC-normalized and counted without special tokens. On NTREX, 95\% intervals come from a bootstrap over its 123 source documents (\texttt{DOCUMENT\_IDS.tsv}, 2{,}000 draws), for premiums and for paired comparisons between two tokenizers; on Universal Dependencies corpora, paired comparisons resample blocks of 10 consecutive sentences. An earlier draft resampled sentences or blocks on NTREX; document-level intervals are about 1.6 times wider, and no conclusion changes. These intervals describe uncertainty within the corpora, not generalization to other text types. No paid inference call was required.

\section{Results}
\subsection{French pays more than Chinese}
\begin{figure}[t]\centering
\includegraphics[width=\linewidth]{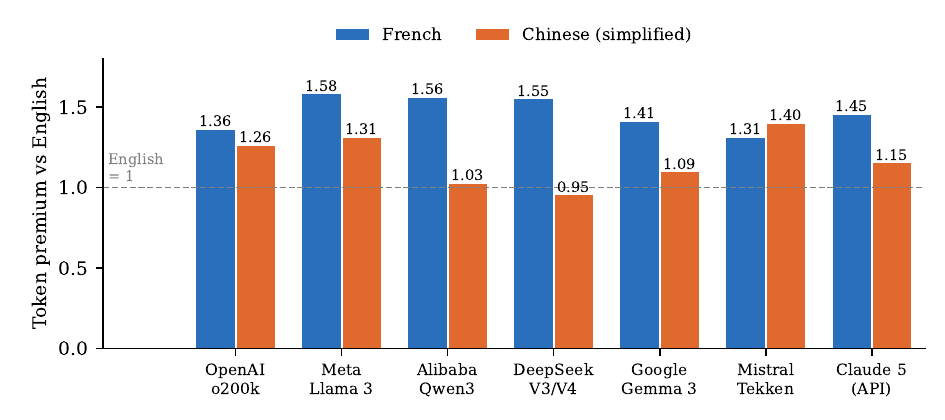}
\caption{French and Chinese token premiums relative to English on NTREX-128, by tokenizer. Publishers: United States (o200k, Llama 3, Gemma 3, Claude), China (Qwen3, DeepSeek), France (Tekken).}\label{fig:frzh}
\end{figure}

Table~\ref{tab:premiums} and Figure~\ref{fig:frzh} give premiums on NTREX. French requires 1.31 (Tekken) to 1.58 (Llama~3) times the English token count, and 1.45 for Claude~5 (95\% CI 1.44--1.47); document-level 95\% intervals never exceed $\pm2.3$ percentage points for French and Chinese. Chinese ranges from 0.95 (DeepSeek) to 1.40 (Tekken) and is cheaper than French on six of seven current tokenizers. Even a high-resource language pays: French is only the 9th to 24th cheapest of the 124 non-English references, while the median language pays 1.69--2.15. The Claude generation-5 tokenizer reduces the French premium relative to Haiku~4.5 (about 1.57 on concatenated text) but segments all texts much more finely: 84{,}000 tokens for the English corpus versus 52{,}000--54{,}000 elsewhere, and 122{,}000 for French versus 71{,}000 for Tekken. Cross-provider price comparisons must therefore be made per document, not per token.

\begin{table}[t]\centering\small
\caption{Token premium relative to English on NTREX-128 (same content). Claude: sentence-level counts via API, French and Chinese only.}\label{tab:premiums}
\begin{tabular}{lccccccc}\toprule
Language & o200k & Llama 3 & Qwen3 & DeepSeek & Gemma 3 & Tekken & Claude 5\\\midrule
French & 1.36 & 1.58 & 1.56 & 1.55 & 1.41 & 1.31 & 1.45\\
Chinese (simpl.) & 1.26 & 1.31 & 1.03 & 0.95 & 1.09 & 1.40 & 1.15\\
Spanish & 1.26 & 1.46 & 1.45 & 1.43 & 1.23 & 1.26 & --\\
German & 1.30 & 1.56 & 1.54 & 1.50 & 1.31 & 1.29 & --\\
Italian & 1.40 & 1.56 & 1.55 & 1.48 & 1.29 & 1.33 & --\\
Arabic & 1.43 & 1.73 & 1.66 & 1.70 & 1.52 & 1.30 & --\\
Hindi & 1.61 & 2.63 & 4.56 & 2.90 & 1.34 & 1.84 & --\\
Basque & 1.61 & 1.88 & 1.89 & 1.75 & 1.67 & 1.54 & --\\
Catalan & 1.45 & 1.67 & 1.66 & 1.59 & 1.50 & 1.42 & --\\
Wolof & 1.62 & 1.78 & 1.79 & 1.75 & 1.64 & 1.73 & --\\
Tahitian & 2.08 & 2.35 & 2.34 & 2.36 & 2.22 & 2.27 & --\\
Khmer & 3.24 & 8.50 & 6.27 & 6.46 & 2.43 & 14.32 & --\\
\bottomrule\end{tabular}\end{table}

\subsection{Regional and overseas languages of France}
\begin{figure}[t]\centering
\includegraphics[width=0.92\linewidth]{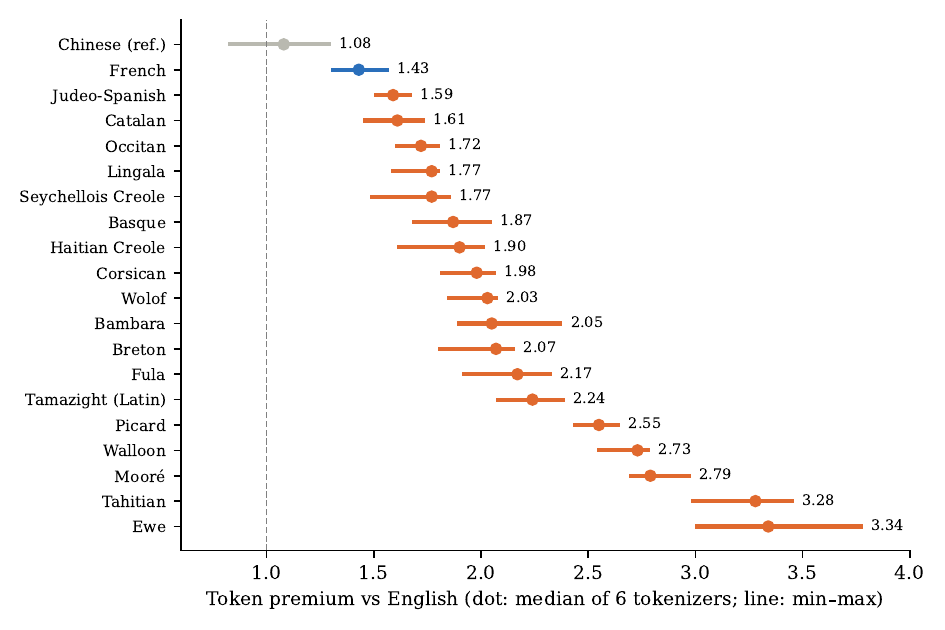}
\caption{Token premium on the UDHR for regional, overseas and Francophone languages; median and range over six 2026 tokenizers.}\label{fig:regional}
\end{figure}

On the UDHR (Figure~\ref{fig:regional}), Breton, Corsican, Wolof and Bambara require about twice the English count; Picard, Walloon and Tahitian 2.5 to 3.5 times. Relative to French itself, the extra cost ranges from +13\% (Catalan) and +20\% (Occitan) to +45\% (Breton), +90\% (Walloon) and +130\% (Tahitian); in Tahitian, the okina written as an apostrophe fragments words heavily. Paragraph structure matches the English version at 95\% or more for 19 of 20 versions (85\% for Wolof); sentence-level correspondence is not verified, so these values are orders of magnitude. Beyond France, non-Latin scripts remain the most penalized on NTREX: Tibetan, Dhivehi, Khmer and Lao reach 10--14$\times$ on the least suited tokenizers, a gap that Gemma~3 reduces two- to threefold.

\subsection{Decomposition}
The French premium factors into the character-length ratio of the translated texts (1.18 on NTREX) and the per-character token ratio, which ranges from 1.11 (Tekken) to 1.33 (Llama~3) and reaches 1.23 for Claude~5 (Table~\ref{tab:decomp}). This decomposition is accounting, not causal: a tokenizer can assign one token to a longer French sequence, and the length ratio is an observation on translated news, not a constant of the language.

\begin{table}[t]\centering\small
\caption{Decomposition of the French premium on NTREX-128.}\label{tab:decomp}
\begin{tabular}{lccc}\toprule
Tokenizer & Characters (fr/en) & Tokens per character (fr/en) & Premium\\\midrule
Baracoda FR v1.2 & 1.18 & 1.01 & 1.20\\
Mistral Tekken & 1.18 & 1.11 & 1.31\\
OpenAI o200k & 1.18 & 1.15 & 1.36\\
Google Gemma 3 & 1.18 & 1.19 & 1.41\\
Claude 5 (API) & 1.18 & 1.23 & 1.45\\
DeepSeek V3/V4 & 1.18 & 1.31 & 1.55\\
Alibaba Qwen3 & 1.18 & 1.32 & 1.56\\
Meta Llama 3 & 1.18 & 1.33 & 1.58\\\bottomrule\end{tabular}\end{table}

\section{Economic implications}
\begin{figure}[t]\centering
\includegraphics[width=\linewidth]{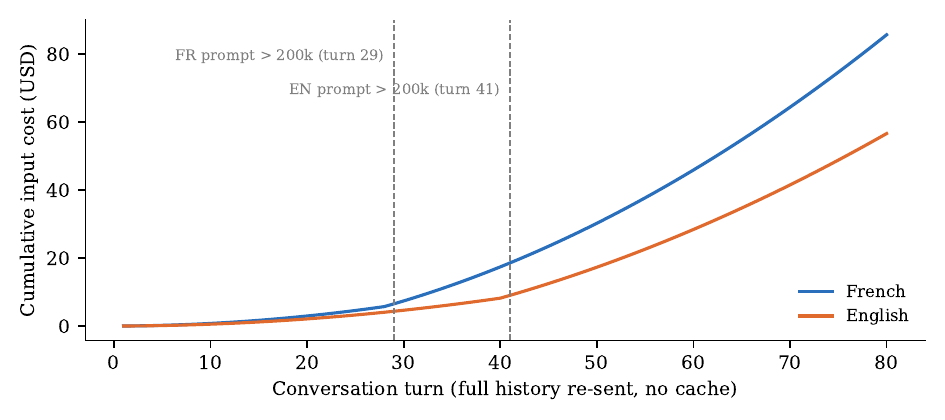}
\caption{Simulated cumulative input cost of an agent loop (5{,}000 English-equivalent tokens added per turn, full history re-sent, Gemini~3.1 Pro tiered prices, French premium 1.41).}\label{fig:long}
\end{figure}

For short requests the surcharge is proportional: at constant per-token prices, input that costs 100 in English costs 131 (Tekken) to 158 (Llama~3) in French. The complete cost of a task also depends on outputs and quality, which we do not measure. In long-context and agentic use, three mechanisms widen the absolute gap. The token ratio per turn stays constant; the price ratio stays constant too, except when a tariff threshold is crossed (ii). (i) When the full history is re-sent at each turn (common, and our scenario, without prompt caching), cumulative cost grows quadratically with the number of turns, and so does the absolute gap. (ii) Tiered pricing is crossed earlier: Google charges Gemini~3.1 Pro input at \$2 per million tokens up to 200k tokens and \$4 beyond~\cite{gcpprice}. In a simulation adding 5{,}000 English-equivalent tokens per turn with the Gemma~3 premium (1.41), the French prompt crosses the threshold at turn 29 and the English one at turn 41; in between each French turn costs 2.8 times the English one, and the cumulative surcharge rises from 41\% to 112\% before receding to 51\% at turn 80 (Figure~\ref{fig:long}). (iii) A 128k-token window holds, in French, the equivalent of 81k--98k tokens of English content (24--37\% less), so it saturates at turn 19 instead of 26 in the same scenario. Attention operations grow quadratically with sequence length in dense attention (1.85--2.5$\times$ for premiums of 1.36--1.58), but this is only one component of total compute, mitigated by other layers, caches and local-attention architectures; we do not infer latency or energy effects from it.

\section{A French-optimized prototype}\label{sec:proto}
\paragraph{Construction.} All versions use byte-level BPE with 131{,}072 tokens (Tekken's size) and a Qwen3-style pre-tokenizer. v1 was trained on 1M Europarl~\cite{europarl} French--English sentence pairs plus 11\,MB of Python; v1.1 changed only pre-tokenization, detaching French elisions (\emph{l'}, \emph{qu'}) and grouping digits by three; v1.2 kept v1.1's settings and changed only the data: 300\,MB of French web text (FineWeb-2), 300\,MB of English web text (FineWeb), 50\,MB of Europarl per language and 36\,MB of Python (736\,MB in total). Any training document containing verbatim a benchmark sentence longer than 40 characters was removed (88 of 794{,}962 documents); paraphrases are not detected. Measured separately on PUD, elisions alone give $-8.9\%$ versus Tekken, digit grouping alone $-10.4\%$, both $-10.6\%$ (v1: $-8.7\%$). Grouping digits may harm arithmetic in downstream models, which remains to be tested.

\paragraph{Benchmark status.} v1 and v1.1 settings were chosen by looking at NTREX and PUD; GSD, Sequoia, Rhapsodie, EWT and GUM were used for post-hoc checks. Although v1.2 settings were fixed before training, all these corpora accompanied the design: we report them as \emph{development and monitoring} benchmarks, and evaluate the frozen tokenizer on a separate final test (Section~\ref{sec:final}). Decoding restores the NFC-normalized text exactly on more than 45{,}000 sentences; this is not byte-level losslessness on arbitrary input: NFD input comes back in NFC form, and a literal \texttt{<|im\_start|>} in the input is parsed as a special token and dropped by default decoding. Counts were reproduced on two machines, from scratch on a fresh environment, and by an independent AI-assisted reproduction from the released files (final-test counts, CroissantLLM and Claude corrections; the later controls in Table~\ref{tab:controls} remain to be reproduced externally).

\begin{table}[t]\centering\small
\caption{Iterations of Baracoda FR: token count relative to Mistral Tekken.}\label{tab:iter}
\begin{tabular}{llcccccc}\toprule
Version & Change & NTREX fr & PUD fr & Rhapsodie (oral fr) & NTREX en & EWT en & Code\\\midrule
v1 & Europarl + code & $-9.9\%$ & $-8.7\%$ & $-6.5\%$ & $+0.4\%$ & $+0.4\%$ & $+1.6\%$\\
v1.1 & pre-tokenization & $-11.3\%$ & $-10.6\%$ & $-6.6\%$ & $-0.7\%$ & $-1.9\%$ & $-1.9\%$\\
v1.2 & web data & $-12.9\%$ & $-12.7\%$ & $-9.8\%$ & $-4.7\%$ & $-5.7\%$ & $-2.7\%$\\\bottomrule\end{tabular}\end{table}

\begin{table}[t]\centering\small
\caption{Baracoda FR v1.2 versus Mistral Tekken (tokens). Intervals: paired block bootstrap, 95\%.}\label{tab:v12}
\begin{tabular}{lrrr}\toprule
Text & v1.2 & Tekken & Difference\\\midrule
NTREX French news & \num{61616} & \num{70724} & $-12.9\%$ [$-13.3$, $-12.5$]\\
UD PUD French & \num{25947} & \num{29707} & $-12.7\%$ [$-13.1$, $-12.2$]\\
UD GSD French (test) & \num{10614} & \num{12226} & $-13.2\%$\\
UDHR French & \num{2272} & \num{2651} & $-14.3\%$\\
UD Rhapsodie (spoken French) & \num{45464} & \num{50384} & $-9.8\%$\\
NTREX English news & \num{51452} & \num{53975} & $-4.7\%$ [$-5.0$, $-4.3$]\\
UD PUD English & \num{22742} & \num{23868} & $-4.7\%$ [$-5.1$, $-4.3$]\\
UD EWT English (web) & \num{279313} & \num{296332} & $-5.7\%$\\
UD GUM English & \num{271059} & \num{282269} & $-4.0\%$\\
Python (NumPy source) & \num{830799} & \num{853479} & $-2.7\%$\\
NTREX Spanish & \num{87920} & \num{67758} & $+29.8\%$\\
NTREX German & \num{96507} & \num{69764} & $+38.3\%$\\
UDHR Breton / Corsican / Occitan & -- & -- & $+3.7\%$ / $+7.1\%$ / $+5.6\%$\\
NTREX Chinese & \num{155652} & \num{75374} & $\times2.1$\\\bottomrule\end{tabular}\end{table}

\begin{figure}[t]\centering
\includegraphics[width=\linewidth]{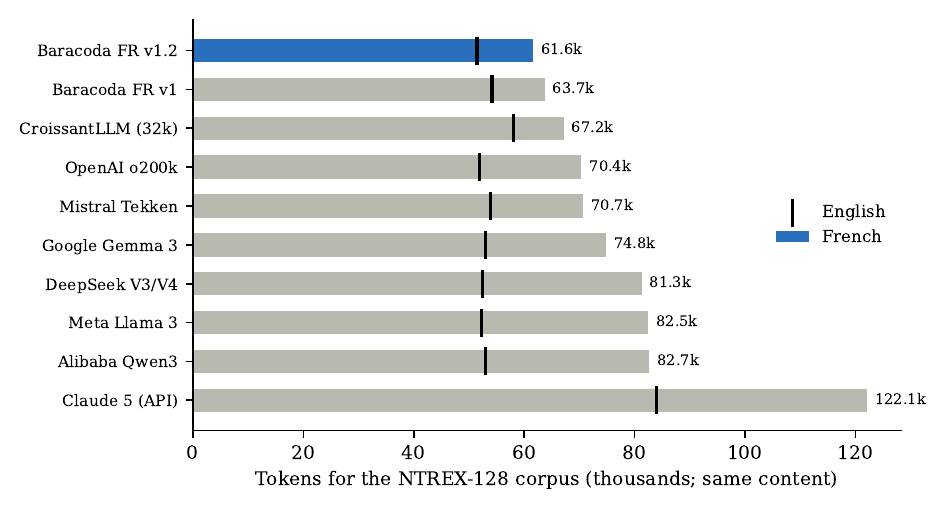}
\caption{Absolute token counts on NTREX-128 (bars: French; ticks: English).}\label{fig:abs}
\end{figure}

\paragraph{Results.} Table~\ref{tab:iter} shows that each change is measured separately; Table~\ref{tab:v12} details v1.2. On NTREX, v1.2 uses fewer tokens than every tokenizer tested, in French and in English (Figure~\ref{fig:abs}), including fewer English tokens than o200k (51{,}452 vs.\ 51{,}860) with a smaller vocabulary. Its French--English premium (1.20 on NTREX) is slightly worse than v1's (1.18) because English improved faster than French, an illustration that parity and efficiency are different objectives.

\begin{figure}[t]\centering
\includegraphics[width=\linewidth]{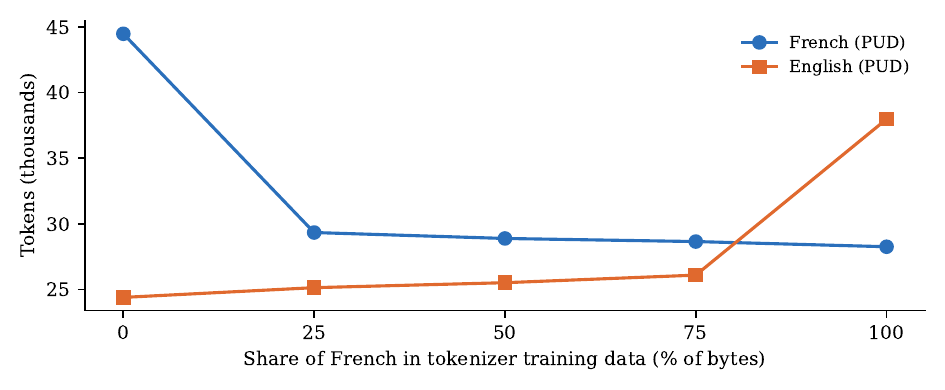}
\caption{Effect of the French share in tokenizer training data (fixed algorithm, 50k vocabulary and 48\,MB volume; mean of three draws), UD PUD.}\label{fig:ablation}
\end{figure}

\paragraph{Controlled experiment.} To isolate the effect of language mix, we trained tokenizers with identical algorithm (BPE, Qwen3 pre-tokenizer), vocabulary (\num{50000}, reached in every run) and data volume (48 million bytes of Europarl French--English, no code), varying only the French share, measured in UTF-8 bytes and made of whole sentences, over three random draws (Figure~\ref{fig:ablation}). Moving from 0\% to 25\% French reduces the PUD premium from 1.81--1.83 to 1.17; beyond that, French gains are marginal ($-2.3\%$ French tokens from 25\% to 75\%) while English degrades ($+3.8\%$), down to a premium of 0.74--0.75 obtained by sacrificing English. Draws differ by less than one point. A modest share of French captures most of the gain, and the premium alone can mislead. An earlier version of this experiment, flagged by an independent AI-assisted reproduction, set the share in characters rather than bytes and targeted \num{131072} tokens while the learned vocabulary only reached 56.8k--87.1k for lack of data; redone in bytes with that target, it gives very close premiums (1.82; 1.15--1.16; 1.13--1.14; 1.11; 0.77). It remains to be replicated on web data and at other data sizes. It does not explain why commercial tokenizers do worse: they cover many languages and code in one vocabulary, and their data are not public.

\paragraph{Comparison with CroissantLLM.} Table~\ref{tab:pud} compares all tokenizers on UD PUD. CroissantLLM has the best parity (1.12 vs.\ 1.14), while v1.2 uses 11.1\% fewer French and 12.4\% fewer English tokens. This advantage is mainly due to vocabulary size. Trained on the same data as v1.2 but with 32{,}768 tokens, our recipe is on par with CroissantLLM on written French ($-0.6\%$ to $-1.3\%$ on UD sets, but $+4.6\%$ on NTREX), slightly better on English ($-0.4\%$ to $-3.3\%$) and clearly worse on spoken French ($+8.9\%$ on Rhapsodie). At comparable size (\num{32768} vs.\ \num{32000} tokens), we do not outperform CroissantLLM, which was trained on a much larger and more varied corpus. On the final test (Table~\ref{tab:final}), v1.2 uses 4.0\% fewer French tokens than CroissantLLM overall but 2.1\% more on spontaneous speech.

\begin{table}[t]\centering\small
\caption{Tokens on UD PUD (1{,}000 parallel sentences) and UD GSD French test.}\label{tab:pud}
\setlength{\tabcolsep}{3.5pt}
\begin{tabular}{lrrrrrrrrr}\toprule
 & Baracoda & Croissant & Tekken & o200k & Gemma 3 & DeepSeek & Llama 3 & Qwen3 & Claude 5\\\midrule
PUD French & \num{25947} & \num{29180} & \num{29707} & \num{29826} & \num{30900} & \num{33695} & \num{35162} & \num{35371} & \num{52059}\\
PUD English & \num{22742} & \num{25963} & \num{23868} & \num{22999} & \num{23389} & \num{23148} & \num{23219} & \num{23750} & \num{36578}\\
Premium & 1.14 & 1.12 & 1.24 & 1.30 & 1.32 & 1.46 & 1.51 & 1.49 & 1.42\\
GSD French & \num{10614} & \num{11766} & \num{12226} & \num{12002} & \num{12609} & \num{13545} & \num{13996} & \num{14124} & --\\\bottomrule\end{tabular}\end{table}

\subsection{Final test with a protocol fixed in advance}\label{sec:final}
Before any computation, we wrote a protocol fixing the frozen tokenizer (SHA-256 published), six Universal Dependencies corpora never consulted during design (French: ParTUT, ParisStories, FQB; English: ParTUT, LinES, GENTLE; all splits), and three hypotheses: H1, fewer tokens than Tekken on each French corpus; H1b, at least 8\% fewer on pooled French; H2, at most 2\% more on pooled English. The SHA-256 digests prove file identity, not date; lacking a prior public deposit, we call the protocol \emph{declared fixed before the test} rather than pre-registered, and will deposit future protocols' digests publicly before computing. All three hold (Table~\ref{tab:final}). Claude~5, counted sentence by sentence on the same corpora, uses \num{188896} French and \num{286135} English tokens, 73\% and 50\% more than Tekken. Two controls follow (Table~\ref{tab:controls}). \emph{Overlap with training data:} of the \num{11198} final-test sentence occurrences longer than 40 characters, 503 (4.49\%; 501 distinct texts) appear verbatim in the v1.2 training data, almost all Europarl excerpts contained in ParTUT; training data had been decontaminated only against the development benchmarks, since the final-test corpora were chosen afterwards. Removing all 503 occurrences leaves the gains essentially unchanged. \emph{Vocabulary budget:} Tekken has \num{131072} entries of which \num{1000} are reserved special tokens (\num{130072} ordinary), whereas v1.2 has \num{131068} ordinary and 4 special tokens; retrained with exactly \num{130072} ordinary tokens, the gains change by 0.012 points in French and 0.026 points in English.

\begin{table}[t]\centering\small
\caption{Final test (tokens), protocol declared fixed before the test. Intervals: paired block bootstrap (blocks of 10 sentences), 95\%.}\label{tab:final}
\setlength{\tabcolsep}{4pt}
\begin{tabular}{lrrrrr}\toprule
Corpus & v1.2 & Tekken & v1.2 vs Tekken & CroissantLLM & o200k\\\midrule
French ParTUT (legal, web, Wikipedia) & \num{28496} & \num{33248} & $-14.3\%$ [$-15.0$, $-13.7$] & \num{30488} & \num{32498}\\
French ParisStories (spontaneous speech) & \num{43281} & \num{47824} & $-9.5\%$ [$-9.8$, $-9.2$] & \num{42376} & \num{47844}\\
French FQB (questions) & \num{25016} & \num{28283} & $-11.6\%$ [$-11.9$, $-11.2$] & \num{27922} & \num{28236}\\
\textbf{French, pooled} & \num{96793} & \num{109355} & $\mathbf{-11.5\%}$ [$-11.9$, $-11.1$] & \num{100786} & \num{108578}\\
English ParTUT & \num{52028} & \num{54670} & $-4.8\%$ [$-5.3$, $-4.4$] & \num{58400} & \num{52847}\\
English LinES & \num{111818} & \num{115331} & $-3.0\%$ [$-3.2$, $-2.9$] & \num{120871} & \num{112343}\\
English GENTLE & \num{20317} & \num{21164} & $-4.0\%$ [$-4.7$, $-3.3$] & \num{22676} & \num{20268}\\
\textbf{English, pooled} & \num{184163} & \num{191165} & $\mathbf{-3.7\%}$ [$-3.9$, $-3.5$] & \num{201947} & \num{185458}\\\bottomrule
\end{tabular}\end{table}

\begin{table}[t]\centering\small
\caption{Controls on the final test: v1.2 vs.\ Tekken after removing sentence occurrences that overlap the v1.2 training data, and with an ordinary-token budget equal to Tekken's. Intervals: paired block bootstrap (blocks of 10 sentences), 95\%.}\label{tab:controls}
\setlength{\tabcolsep}{3.5pt}
\resizebox{\linewidth}{!}{\begin{tabular}{lrrrrrrl}\toprule
Corpus & Sentences & $>$40 chars & Overlap & Retained & v1.2 & Tekken & v1.2 vs Tekken\\\midrule
French ParTUT & \num{1020} & 948 & 175 & 845 & \num{23636} & \num{27698} & $-14.67\%$ [$-15.39$, $-13.95$]\\
French ParisStories & \num{2776} & \num{1673} & 0 & \num{2776} & \num{43281} & \num{47824} & $-9.50\%$ [$-9.81$, $-9.16$]\\
French FQB & \num{2289} & \num{1470} & 6 & \num{2283} & \num{24950} & \num{28208} & $-11.55\%$ [$-11.93$, $-11.15$]\\
\textbf{French, pooled} & \num{6085} & \num{4091} & 181 & \num{5904} & \num{91867} & \num{103730} & $\mathbf{-11.44\%}$ [$-11.82$, $-11.08$]\\
English ParTUT & \num{2090} & \num{1928} & 314 & \num{1776} & \num{43037} & \num{45348} & $-5.10\%$ [$-5.65$, $-4.60$]\\
English LinES & \num{5696} & \num{4469} & 8 & \num{5688} & \num{111728} & \num{115235} & $-3.04\%$ [$-3.21$, $-2.88$]\\
English GENTLE & \num{1334} & 710 & 0 & \num{1334} & \num{20317} & \num{21164} & $-4.00\%$ [$-4.71$, $-3.34$]\\
\textbf{English, pooled} & \num{9120} & \num{7107} & 322 & \num{8798} & \num{175082} & \num{181747} & $\mathbf{-3.67\%}$ [$-3.88$, $-3.47$]\\\midrule
\multicolumn{8}{l}{\emph{Equal ordinary-token budget} (\num{130072} ordinary + 4 special tokens; v1.2: \num{131068} + 4; Tekken: \num{130072} + \num{1000})}\\
French, pooled, all sentences & \num{6085} & & & & \num{96806} & \num{109355} & $-11.475\%$ [$-11.85$, $-11.14$] (v1.2: $-11.487\%$)\\
English, pooled, all sentences & \num{9120} & & & & \num{184212} & \num{191165} & $-3.637\%$ [$-3.83$, $-3.45$] (v1.2: $-3.663\%$)\\\bottomrule
\end{tabular}}\end{table}

\section{Limitations}
We measure tokens, not invoices, model quality or task success. A tokenizer is not a model: using Baracoda~FR requires retraining at least the embeddings of a model, with a compute cost and a quality risk to be measured; extending the Qwen3 tokenizer with 24{,}269 French tokens only moved its premium from 1.56 to 1.43 in our simple implementation. The prototype is a specialization trade-off, not a general reduction of linguistic inequality. Apart from the final test, the benchmarks used for the prototype are development benchmarks; the final test covers French and English only. Claude counts rely on a message overhead measured by difference (7 tokens) and on an endpoint that Anthropic describes as an estimate. The UDHR is a single document per language; each version's structure matches the English one at 95\% or more (85\% for Wolof, whose paragraphing differs), but sentence-level correspondence is not verified. Comparisons involve vocabularies from 32k to 262k tokens.

\section{Conclusion}
In 2026, working in French with an LLM still costs 31\% to 58\% more tokens than in English for the same content, and regional languages of France pay about double. The gap is not a property of the language alone: a tokenizer trained with enough French narrows it substantially, as our prototype shows on segmentation benchmarks, including a final test with a protocol fixed in advance. The next step is to show that such a tokenizer reduces the cost of correctly completed tasks in a model, starting with banking and insurance French, with a control branch, established vocabulary transfer methods~\cite{focus2023,tokalign2025}, cross-word tokens~\cite{superbpe2025} and a final test whose protocol digest is deposited publicly beforehand.

\paragraph{Data and code availability.} Tokenizer files, scripts, per-sentence counts (including Claude's), controls and library versions are released at \url{https://github.com/Baracoda-ai-labs/baracoda-fr} (version cited: commit \texttt{fc8a736}, 29 September 2026): code and tokenizer files under the Apache License~2.0, results and documentation under CC~BY~4.0. Training data and evaluation corpora are not redistributed; scripts download them at pinned versions.

\paragraph{Acknowledgements.} Measurements, analyses and drafting were carried out with the assistance of AI systems (Anthropic's Claude); an independent AI-assisted review and reproduction of earlier versions (not a certification by an independent laboratory) shaped the protocol and the controls. All figures were regenerated from the files of the accompanying reproduction pack.

\end{document}